\documentclass[runningheads]{llncs}
\usepackage[T1]{fontenc}
\usepackage{graphicx}
\usepackage{booktabs}
\usepackage[misc]{ifsym}
\newcommand{\corr}{(\Letter)}
\usepackage{mwe}

\usepackage{hyperref}
\usepackage{amssymb}
\usepackage{mathtools}
\usepackage{tabularx}
\usepackage{booktabs}
\usepackage{multirow}
\usepackage{algorithm}
\usepackage{algorithmic}
\usepackage{rotating}
\usepackage{subcaption}
\usepackage[table]{xcolor}
\usepackage{xspace}
\usepackage{graphicx}
\usepackage{authblk}

\begin{document}
\title{ST-LoRA: Low-rank Adaptation for Spatio-Temporal Forecasting}
\toctitle{ST-LoRA: Low-rank Adaptation for Spatio-Temporal Forecasting}
\titlerunning{ST-LoRA: Low-rank Adaptation for Spatio-Temporal Forecasting}

\newcommand\relatedversion{}
\renewcommand\relatedversion{\thanks{The full version of the paper can be accessed at \protect\url{https://arxiv.org/abs/2404.07919}}} 

\author{Weilin Ruan\inst{1} \and
Wei Chen\inst{1} \and
Xilin Dang\inst{2} \and
Jianxiang Zhou\inst{1} \and
Weichuang Li\inst{1} \and \\
Xu Liu\inst{3} \and
Yuxuan Liang\inst{1,4} \corr 
}

\tocauthor{Weilin Ruan, Wei Chen, Xilin Dang, Jianxiang Zhou, Weichuang Li, Xu Liu, Yuxuan Liang}


\authorrunning{Ruan et al.}

\institute{The Hong Kong University of Science and Technology (Guangzhou), China 
\email{\{rwlinno,onedeanxxx\}@gmail.com}, \email{\{jzhou814,wli043\}@connect.hkust-gz.edu.cn}, \email{yuxliang@outlook.com}
\and
The Chinese University of Hong Kong, Hong Kong \email{xldang23@cse.cuhk.edu.hk}
\and
National University of Singapore, Singapore \email{liuxu@comp.nus.edu.sg}
\and
State Key Lab of Resources and Environmental Information System, Chinese Academy of Sciences, China \email{yuxliang@outlook.com}
}

\maketitle 

\def\model{ST-LoRA\xspace}

\begin{abstract}
Spatio-temporal forecasting is essential for understanding future dynamics within real-world systems by leveraging historical data from multiple locations. Existing methods often prioritize the development of intricate neural networks to capture the complex dependencies of the data. These methods neglect node-level heterogeneity and face over-parameterization when attempting to model node-specific characteristics. In this paper, we present a novel \underline{lo}w-\underline{r}ank \underline{a}daptation framework for existing \underline{s}patio-\underline{t}emporal prediction models, termed \model, which alleviates the aforementioned problems through node-level adjustments. Specifically, we introduce the node-adaptive low-rank layer and node-specific predictor, capturing the complex functional characteristics of nodes while maintaining computational efficiency. Extensive experiments on multiple real-world datasets demonstrate that our method consistently achieves superior performance across various forecasting models with minimal computational overhead, improving performance by 7\% with only 1\% additional parameter cost. The source code is available at \url{https://github.com/RWLinno/ST-LoRA}.
\end{abstract}

\section{Introduction}
With the rapid advancement of data acquisition technologies and mobile computing, vast spatio-temporal data are being generated for urban analysis and related applications~\cite{zhang2011data,du2018sensable}. Spatio-temporal forecasting aims to predict future changes based on dynamic temporal observations recorded at static locations with spatial associations~\cite{zhang2020spatio}. Modeling and analyzing these spatio-temporal dynamic systems can be applied to various prediction scenarios, such as traffic speed forecasting~\cite{yu2018spatio,wu2019graph}, taxi demand prediction~\cite{yao2018deep}, and air quality prediction~\cite{liang2018geoman,liang2023airformer}.
Early research primarily focused on traditional time-series models, such as the Historical Average (HA)~\cite{smith1997traffic} method and the Auto-Regressive Integrated Moving Average (ARIMA)~\cite{lippi2013short} model, as well as machine learning-based models~\cite{van2012short}, including Vector Auto-Regression (VAR)~\cite{zivot2006vector} and Artificial Neural Networks (ANN)~\cite{huang2014deep}. These methods were applied directly to spatio-temporal forecasting (STF) without considering spatial dependencies, leading to suboptimal performance. With the accumulation of spatio-temporal big data, recent approaches have shifted towards data-driven deep learning models, which are capable of capturing the inherent spatio-temporal dependencies within dynamic systems. Simple yet effective strategies involve using convolutional neural networks (CNNs)~\cite{gu2018recent} to capture spatial dependencies and recurrent neural networks (RNNs)~\cite{yu2019review,hochreiter1997long,zhang2018combining,chung2014empirical} for temporal dependencies, thereby improving performance.
\begin{figure}[t]
    \centering
    \includegraphics[width=\linewidth]{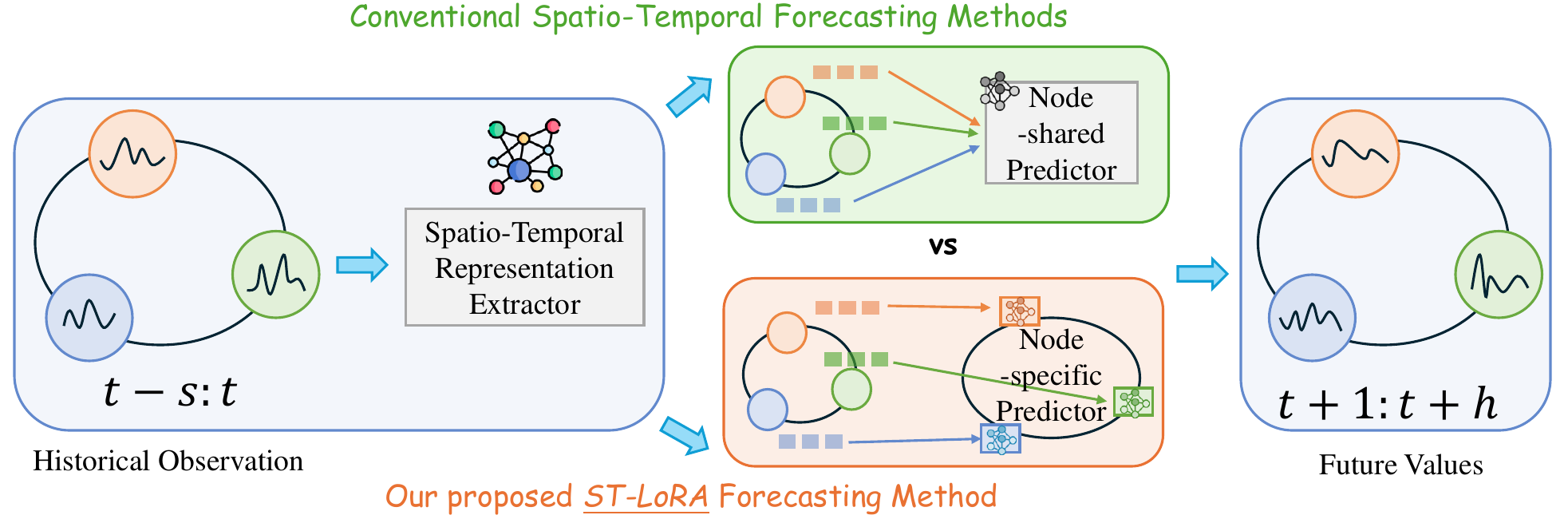}
    \caption{Comparison between conventional STF methods and our proposed ST-LoRA framework featuring node-specific predictors for node-level fine-tuning.}
    \label{fig:compare}
\end{figure}

Given the non-Euclidean nature of spatial dependencies, deep learning methods have evolved to combine sophisticated temporal models with Graph Neural Networks (GNNs)~\cite{wu2020comprehensive,kipf2016semi} for capturing both global temporal dependencies and regional patterns. Spatio-temporal graph neural networks (STGNNs)~\cite{li2017diffusion,yu2018spatio} have emerged as powerful tools for learning robust high-level spatio-temporal representations through local information aggregation~\cite{jin2023spatio}. Recent years have witnessed significant advances in this field, including innovations in graph convolution architectures~\cite{fang2021spatial,wang2020traffic}, dynamic graph structure learning~\cite{wu2019graph,jiang2023spatio}, and efficient attention mechanisms~\cite{zhou2022fedformer,wu2021autoformer}. Furthermore, researchers have explored integrating advanced techniques such as self-supervised learning~\cite{shao2022pre} and large language models~\cite{zhou2024one} into spatio-temporal prediction tasks. While these sophisticated approaches have achieved remarkable performance improvements, they often come at the cost of increased computational complexity and memory requirements, making it challenging to balance model effectiveness with operational efficiency.

To better understand the landscape of spatio-temporal forecasting, we present a systematic analysis of existing architectures, revealing their common structural patterns as illustrated in Figure~\ref{fig:compare}. Existing spatio-temporal forecasting methods typically consist of two main components. The first component is the \textit{Spatio-Temporal Representation Extractor}, which serves as the core framework and is responsible for capturing high-order complex spatio-temporal relationships. This component can be implemented using various architectures, such as CNNs, RNNs, and STGNNs. The second component is the \textit{Node-Shared Predictor}, which takes the advanced spatio-temporal representations extracted by the first component and predicts future changes for each location. This predictor typically consists of parameter-sharing fully connected layers.
\begin{figure}[!t]
    \centering
    \includegraphics[width=\linewidth]{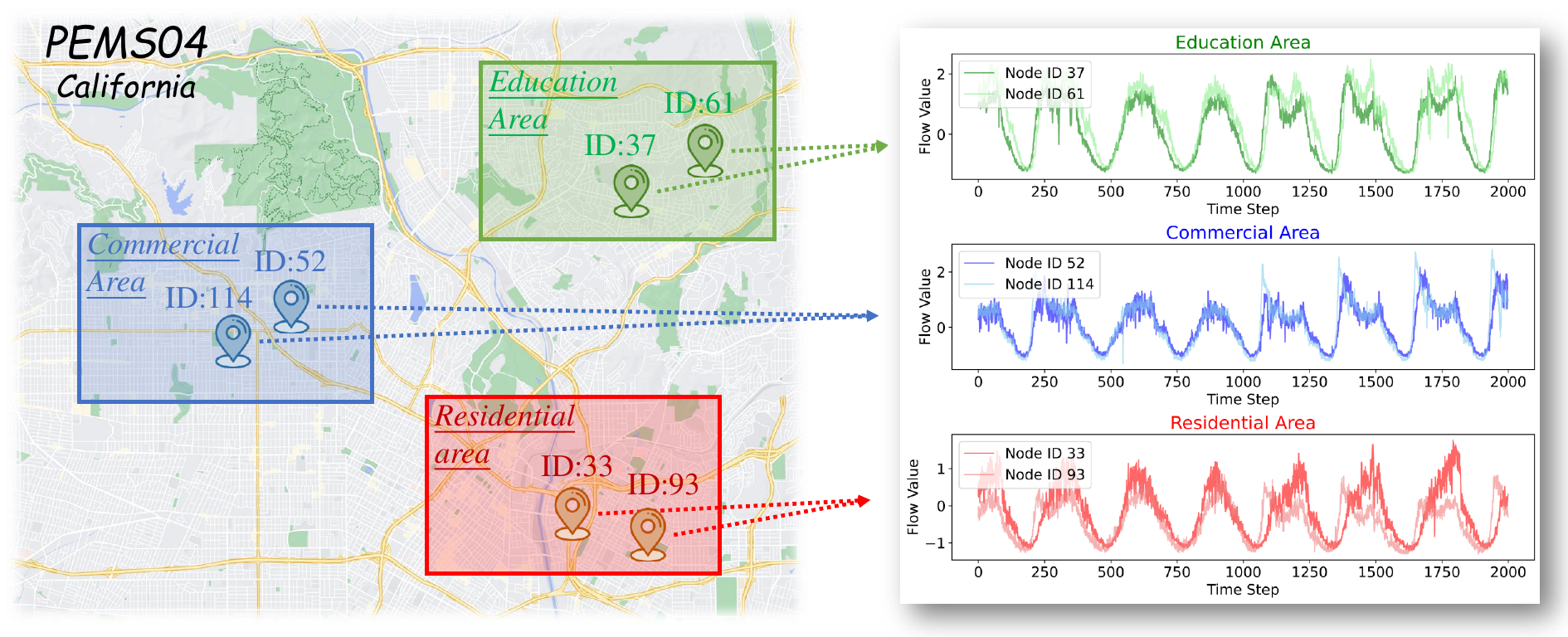}
    \caption{Traffic flow visualization of each node in different areas on the PEMS04.}
    \label{fig:motivation}
\end{figure}

While this two-component architecture has been widely adopted in existing methods, it suffers from a significant limitation: the parameter-sharing node predictor fundamentally struggles to address \textbf{node-level heterogeneity}. This shared parameterization approach assumes all nodes exhibit similar behavior patterns that can be modeled using identical parameters. However, in real-world scenarios, nodes frequently display distinct temporal dynamics and behavioral characteristics that shared predictors cannot adequately capture. This heterogeneity is particularly pronounced in urban traffic networks, where sensors distributed across diverse functional areas exhibit unique patterns influenced by their specific contexts and surrounding environments. To illustrate this problem intuitively, we analyze the PEMS04 traffic flow dataset~\cite{chen2001freeway} from California, examining nodes across different urban zones (Figure~\ref{fig:motivation}). Our analysis reveals that even nodes 33 and 93, situated within the same residential area, exhibit substantially different temporal evolution patterns between time steps 1250 and 2000. This finding underscores the limitations of current parameter-sharing approaches in modeling node-specific behavioral characteristics. Consequently, existing spatio-temporal forecasting models attempting to address heterogeneity through extensive shared parameterization inevitably confront the \textbf{over-parameterization dilemma}. Maintaining separate parameters for each node not only incurs prohibitive computational and memory costs, especially for large-scale networks with hundreds or thousands of nodes, but also significantly increases the risk of model overfitting and compromised generalization performance.

To address these challenges, we draw inspiration from low-rank matrix factorization techniques~\cite{sainath2013low}. Specifically, we first customize a node-adaptive low-rank layer containing multiple trainable matrices, utilizing low-rank decomposition techniques to effectively reduce computational complexity and enhance model training efficiency. Subsequently, we propose a novel lightweight and efficient low-rank adaptation framework named \model. This framework seamlessly integrates the low-rank layer into existing spatio-temporal forecasting models through a multi-layer fusion residual stacking approach, thereby achieving node-specific predictors and mitigating the effects of overparameterization. Experimental results on real-world traffic datasets show that our framework significantly improves performance over various baseline methods in spatio-temporal forecasting tasks. Our major contributions can be summarized as follows:

\noindent \textbf{1) A node-level heterogeneity perspective for STF.} We are the first to introduce low-rank adaptation techniques to the spatio-temporal domain to explicitly account for the node-level heterogeneity. Our proposed Node Adaptive Low-rank Layers capture diverse node-level patterns and distributions by leveraging low-rank matrix factorization while maintaining computational efficiency.

\noindent \textbf{2) A general low-rank adaptation method for existing ST models.} We developed node-specific predictors along with a framework called \model, which, in a lightweight and efficient manner, allows existing spatio-temporal prediction models to serve as backbone networks to enhance overall performance.

\noindent \textbf{3) Extensive empirical studies.} We rigorously evaluate our proposed method on various models and six public traffic datasets. The experimental results demonstrate that our method significantly enhances prediction accuracy across all baseline models while requiring less than \textbf{1\%} additional learnable parameters, achieving remarkable more than \textbf{7\%} performance improvements in terms of average RMSE across prediction horizons.

\section{Preliminaries}
\subsection{Formulation}
The objective of STF is to predict future values based on previously observed time series data from $N$ correlated sensors. This sensor network can be represented as a weighted directed graph $\mathcal{G} = (\mathcal{V}, \mathcal{E}, \mathcal{W})$, where $\mathcal{V}$ is the node set with $|\mathcal{V}|=N$, $\mathcal{E}$ is the edge set, and $\mathcal{W}\in \mathbb{R}^{N \times N}$ is a weighted adjacency matrix that encodes the relationships between nodes. The spatio-temporal data observed on $\mathcal{G}$ can be represented as a graph signal $X\in \mathbb{R}^{N \times F}$, where $F$ is the number of features associated with each node. Let $X^{(t)}$ denote the graph signal observed at time $t$. The spatio-temporal forecasting problem aims to learn a function $\mathcal{F}(\cdot)$ that maps $s$ historical graph signals to $h$ future graph signals, given a graph $\mathcal{G}$:
\begin{equation}
[X^{(t-s + 1)}, \ldots, X^{(t)}; \mathcal{G}] \stackrel{\mathcal{F}(\cdot)}{\longrightarrow} [X^{(t+1)}, \ldots, X^{(t+h)}].
\end{equation}

\subsection{Related Work}
\subsubsection{Spatio-temporal Forecasting} has evolved into a foundational paradigm for predicting future states by leveraging historical observations across spatial and temporal dimensions. Traditional methods grounded in statistical and time series analysis achieved modest success but exhibited significant limitations in modeling complex spatial structures and intricate ST relationships~\cite{wang2020traffic,jin2023spatio}. To address these shortcomings, deep learning frameworks have increasingly been embraced, which demonstrate superior capability in extracting latent feature representations, including non-linear spatial and temporal correlations from historical data~\cite{wang2020deep,lv2014traffic,wang2022hierarchical}. Among these advanced frameworks, Spatio-Temporal Graph Neural Networks (STGNNs) have emerged as powerful tools for prediction tasks. By integrating Graph Neural Networks (GNNs)~\cite{kipf2016semi} with sophisticated temporal modeling techniques~\cite{yu2019review}, these architectures effectively capture complex spatio-temporal dynamics through local information aggregation. Over the past decade, several influential STGNN architectures have been proposed, including GWNet~\cite{wu2019graph}, STGCN~\cite{yu2018spatio}, DCRNN~\cite{li2017diffusion}, and AGCRN~\cite{bai2020adaptive}, each demonstrating remarkable performance across diverse spatio-temporal prediction tasks. Complementing these developments, attention mechanisms~\cite{vaswani2017attention,ma2024spatio} have gained substantial traction due to their effectiveness in modeling dynamic dependencies inherent in spatio-temporal data. Despite the proliferation and diversification of STGNN architectures, performance improvements have begun to plateau, prompting researchers to explore integrating Self-Supervised Learning (SSL)~\cite{shao2022pre,li2022spatial} and Large Language Models (LLMs)~\cite{zhou2024one,jin2024position,yan2024urbanclip}. Recent studies have further investigated methods to capture spatio-temporal heterogeneity through techniques such as spatial-temporal decoupled masked pre-training~\cite{gao2023spatial} and heterogeneity-informed learning approaches~\cite{dong2024heterogeneity}.

However, these methods often introduce substantial computational overhead and model complexity. Against this backdrop, we introduce a parameter-efficient node-specific adaptation method that significantly enhances existing forecasting frameworks while maintaining minimal parameter and computational costs.

\subsubsection{Low-rank Adaptation} is the technique that decomposes high-dimensional parameter spaces into products of low-rank matrices, reducing computational complexity while preserving essential information. The foundational work on LoRA~\cite{hu2021lora,hayou2024lora+} demonstrated that injecting trainable low-rank matrices into pre-trained models enables efficient adaptation with minimal parameter overhead. This approach has been refined through variants like DyLoRA~\cite{valipour2022dylora} with dynamic adaptation mechanisms and Compacter~\cite{karimi2021compacter} leveraging parameterized complex multiplication layers for task-specific optimization. The extension of low-rank adaptation to multi-modal and spatio-temporal domains represents a significant advancement for complex applications. MTLoRA~\cite{agiza2024mtlora} adapted this approach for multi-task learning scenarios, while robust low-rank reconstruction techniques~\cite{jhuo2012robust} have shown effectiveness for preserving invariant features across domains. In spatio-temporal forecasting, MFSTN~\cite{pan2019matrix} and DeepLGN~\cite{liang2021revisiting} use matrix or tensor factorization for region-specific parameter decomposition, but their grid-based modeling limits node-level granularity. ST-Adapter~\cite{pan2022st} introduced low-rank adaptation for cross-modality transfer in spatio-temporal tasks but focused on modality transfer rather than node-specific adaptation.

However, these approaches rely predominantly on region-level predictions, limiting their ability to capture fine-grained node-level heterogeneity and compromising generalizability across diverse spatial configurations. Our proposed framework addresses these limitations by adapting low-rank optimization for node-level heterogeneity in spatio-temporal forecasting.
\begin{figure}[!t]
    \centering
    \includegraphics[width=\linewidth]{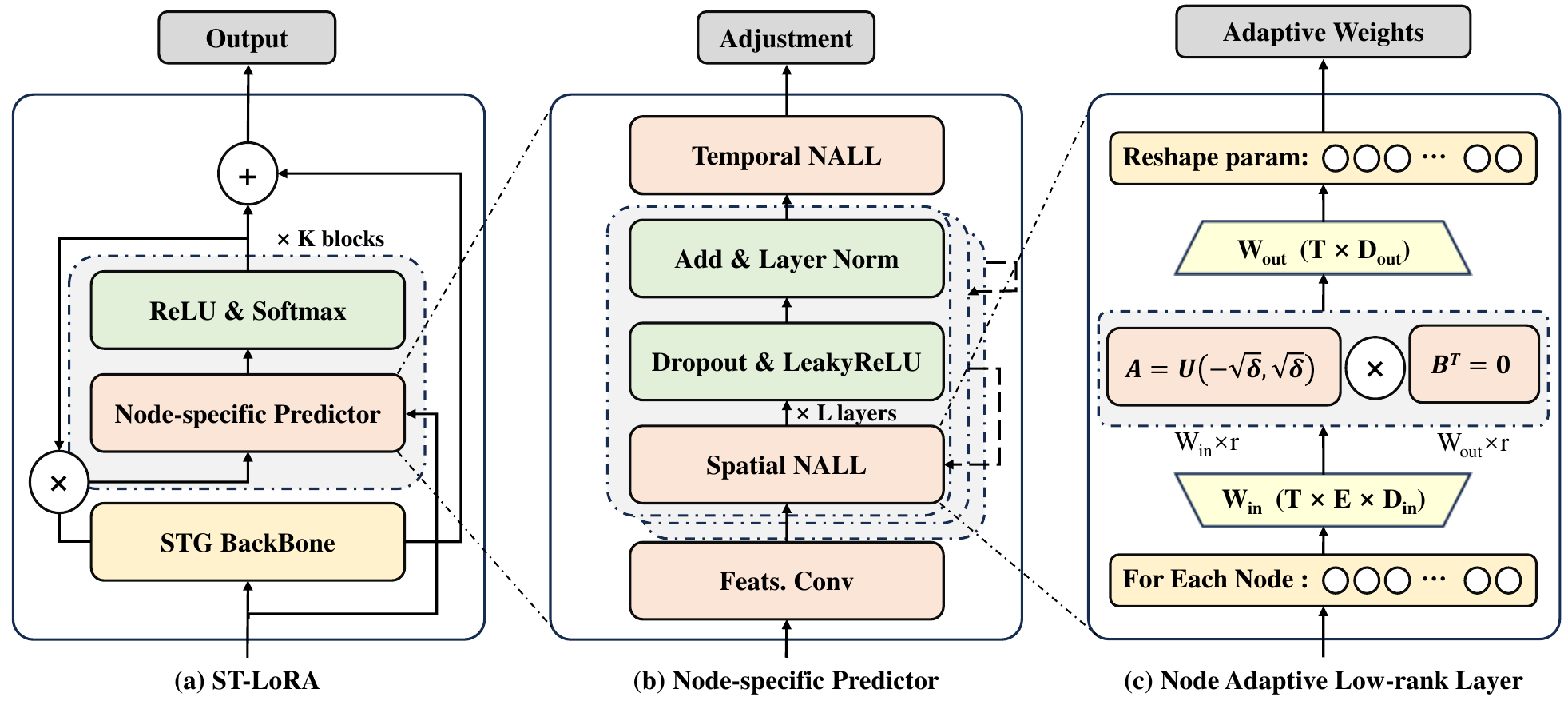}
    \caption{Overview of the proposed \model framework.}
    \label{fig:overview}
\end{figure}

\section{Methodology}
In this section, we present \model, a lightweight yet effective framework that enhances spatio-temporal models by addressing the node heterogeneity and overparameterization challenges. As illustrated in Figure~\ref{fig:overview}, our framework consists of three key components: (a) the gated integration mechanism that enhances existing backbone STGNNs, (b) Node-specific Predictors (NSPs) that capture patterns with node-level fine-tuning, and (c) Node-Adaptive Low-rank Layers (NALLs) that leverage low-rank adaptation for parameter customization.

\subsection{Node-Adaptive Low-rank Layers}
The core innovation of \model lies in the Node-Adaptive Low-rank Layers (NALL), which introduce learnable low-rank matrices to efficiently customize the base model parameters for different nodes. Let $\mathbf{W} \in \mathbb{R}^{d_{out} \times d_{in}}$ represent the base weight matrix, where $d_{in}$ and $d_{out}$ are the input and output dimensions, respectively. Traditional approaches would require a complete set of parameters for each node, leading to $\mathcal{O}(N \cdot d_{in} \cdot d_{out})$ parameters for $N$ nodes, which is computationally prohibitive for large-scale spatio-temporal networks. Drawing inspiration from low-rank matrix factorization theory~\cite{li2018algorithmic}, NALL decomposes the parameter matrix $\Delta \mathbf{W}$ into the product of two low-rank matrices. For node $v_i$ with input $\mathbf{x}$, the adaptation process can be formally expressed as:
\begin{equation}
\Delta \mathbf{W}_{v_{i}} = \mathbf{B}\mathbf{A}_{v_{i}} \cdot \frac{\alpha}{r}, \quad
\mathbf{\hat{y}_i} = \sigma\left(\mathbf{W}\mathbf{x} + \Delta \mathbf{W}_{v_{i}}\mathbf{x} + \mathbf{b}\right),
\label{eq:nall}
\end{equation}
where $\mathbf{B} \in \mathbb{R}^{d_{out} \times r}$ and $\mathbf{A}_i \in \mathbb{R}^{r \times d_{in}}$ are learnable low-rank matrices with rank $r$, $\alpha$ is a scaling factor controlling the adaptation magnitude, and $\sigma$ is a non-linear activation function (e.g., LeakyReLU). This formulation reduces the parameter complexity from $\mathcal{O}(N \cdot d_{in} \cdot d_{out})$ to $\mathcal{O}(r \cdot (d_{in} + d_{out}) + N \cdot r^2)$, representing a significant reduction when $r \ll \min(d_{in}, d_{out})$. By constraining adaptations to lower-dimensional node-specific patterns, NALL efficiently captures meaningful behavioral differences while preventing overfitting to node-specific noise patterns. During inference, while $\mathbf{W}$ remains frozen to preserve the pre-trained knowledge, $\mathbf{A}_i$ and $\mathbf{B}$ continue to adapt and capture node-specific patterns. Our implementation further enhances NALL with dropout with $p=0.3$ and Kaiming initialization to stabilize training dynamics and improve convergence stability.

\subsection{Node-specific Predictors}
While a single NALL layer provides efficient parameter adaptation, capturing complex spatio-temporal dependencies requires a more sophisticated architectural design. We propose Node-Specific Predictors (NSP) that hierarchically stack multiple NALL layers with residual connections to model heterogeneous patterns across different nodes. Given an input sequence $\mathbf{X}_{t-T:t} \in \mathbb{R}^{T \times N \times D}$, where $T$ is the sequence length, $N$ is the number of nodes, and $D$ is the feature dimension, NSP processes the input through following integrated components:
\begin{equation}
\begin{aligned}
\mathbf{H}^{(0)} &= \text{Conv2D}(\mathbf{X}_{t-T:t}), \\
\mathbf{H}^{(l)} &= \mathbf{H}^{(l-1)} + \text{NALL}^{(l)}(\sigma(\mathbf{H}^{(l-1)})), \quad l = 1,2,...,L \\
\hat{\mathbf{Y}}_t &= \mathcal{G}_t(\mathbf{H}^{(L)}),
\end{aligned}
\label{eq:nsp}
\end{equation}
where $\mathbf{H}^{(l)} \in \mathbb{R}^{T \times N \times D}$ represents the hidden feature representations at layer $l$, $\text{Conv2D}(\cdot)$ first extracts temporal features using convolution operations with optimized kernel configurations, multiple NALL layers then process spatial information through residual connections, and finally $\mathcal{G}_t$ projects the features for temporal prediction. The activation function $\sigma$ incorporates RMSNorm and dropout to control model complexity and enhance training stability.

The key innovation of NSP lies in its unified approach to spatio-temporal modeling. Unlike traditional GNNs that rely on fixed graph convolutions, NSP leverages the low-rank structure of NALLs to achieve both spatial and temporal adaptability. Specifically, the spatial patterns are captured through the rank-constrained weight matrices in NALLs, which naturally aggregate node-specific information, while the temporal dependencies are modeled through the sequential application of adapted transformations. This design enables NSP to efficiently capture complex node-specific patterns while maintaining computational efficiency through the low-rank structure of NALL layers. The effectiveness of this architecture is particularly evident in scenarios with heterogeneous node behaviors, where traditional fixed-parameter approaches often struggle to capture diverse patterns simultaneously. Importantly, while NSP provides node-specific adaptations, it preserves spatial dependencies captured by the backbone model by operating on representations that already encode inter-node relationships, thus complementing rather than replacing existing spatial modeling mechanisms.

\subsection{\model Integration}
To enhance model generalization and adaptability to diverse spatio-temporal patterns, we propose ST-LoRA, a novel low-rank adaptation framework for spatio-temporal data that seamlessly integrates existing backbone models with our NSP module. Let $f_{\theta}$ denote a pre-trained spatio-temporal backbone model and $\mathcal{H}(\cdot)$ represent our NSP-based enhancement operator. The \model framework employs a hierarchical architecture with multiple NSP blocks to capture complex spatio-temporal dependencies at different scales. Given input sequence $\mathbf{X}_{t-T:t}$, the integration process can be formulated as:
\begin{equation}
\begin{aligned}
\mathbf{Y}_{\text{base}} &= f_{\theta}(\mathbf{X}_{t-T:t}), \\
\mathbf{Z}^{(1)} &= \mathcal{H}^{(1)}(\sigma(\mathbf{Y}_{\text{base}})), \\
\mathbf{Z}^{(k)} &= \mathcal{H}^{(k)}(\sigma([\mathbf{X}_{t-T:t}, \mathbf{Z}^{(k-1)}])), \quad k = 2,\ldots,K\\
\mathbf{R} &= \sigma(\mathcal{F}([\mathbf{X}_{t-T:t}, \frac{1}{K}\sum_{k=1}^K \mathbf{Z}^{(k)}])), \\
\mathbf{Y}_{\text{final}} &= \mathbf{R} \odot \mathbf{Y}_{\text{base}} + (1-\mathbf{R}) \odot \frac{1}{K}\sum_{k=1}^K \mathbf{Z}^{(k)},
\end{aligned}
\label{eq:integration}
\end{equation}

where $K$ is the number of NSP blocks, $\sigma$ is a non-linear activation function (e.g., ReLU), $[\cdot,\cdot]$ denotes feature concatenation, $\mathcal{F}$ is a learnable fusion layer that generates adaptive blending weights, and $\mathbf{R} \in [0,1]$ is a node-specific gating tensor that controls the contribution of the adaptation mechanism. This allows NSP blocks to access both original historical context and refined node-specific representations, enabling the framework to capture diverse spatio-temporal patterns at different scales while maintaining stability through averaging.

The framework is optimized end-to-end using a temporal Mean Absolute Error (MAE) loss with L2 regularization over the prediction horizon $T'$:
\begin{equation}
\mathcal{L} = \frac{1}{T'}\sum_{i=1}^{T'} \|\mathbf{X}_{t+i} - \mathbf{Y}_{\text{final}}^{(t+i)}\|_1 + \lambda\|\alpha\|_2,
\label{eq:loss}
\end{equation}
where $\lambda$ controls the strength of the regularization term on the gating parameter. This formulation ensures accurate predictions through the MAE term while preventing over-reliance on either the backbone model or the adaptation.

The effectiveness of \model stems from its ability to preserve the backbone model's general prediction capability while introducing node-specific adaptations through the NSP blocks, making it suitable for complex real-world scenarios with heterogeneous spatio-temporal patterns. By integrating NALLs and NSPs, ST-LoRA provides an efficient solution to the fundamental trade-off between modeling node-level heterogeneity and managing computational complexity.

\section{Experiments}
In this section, we conduct extensive experiments to investigate the following Research Questions (RQ):
\begin{itemize}
\item \textbf{RQ1:} Can ST-LoRA be seamlessly integrated with various spatio-temporal prediction models?
\item \textbf{RQ2:} How effectively does our framework improve prediction performance across different scenarios?
\item \textbf{RQ3:} What are the computational overhead and parameter costs of our framework?
\item \textbf{RQ4:} How do different architectural choices affect the model's performance?
\end{itemize}
\subsection{Experimental Setup}
\subsubsection{Datasets.}
We evaluate our approach on six public traffic datasets that are widely used in spatio-temporal forecasting research. As summarized in Table~\ref{tab:dataset}, these datasets encompass both traffic speed measurements (METR-LA, PEMS-BAY) and traffic flow records (PEMS03/04/07/08), featuring diverse spatial scales ranging from 170 to 883 sensor nodes and temporal ranges spanning 16,992 to 52,116 timestamps. Following standard practice, we split each dataset chronologically into training, validation, and testing sets with ratios of 7:1:2 for speed datasets and 6:2:2 for flow datasets. All above datasets are divided along the time axis into three non-overlapping parts, including training, validation, and test sets. METR-LA and PEMS-BAY are divided in a fraction of 7:1:2 while PEMS03, PEMS04, PEMS07, and PEMS08 are divided in a fraction of 6:2:2.
\begin{table}[!b]
\centering
\caption{Statistics and description of datasets we used.}
\begin{tabular}{c|cccccc}
\toprule
Dataset & \#Nodes & \#Edges & \#Frames & Time Range & Type\\ 
\hline
METR-LA & 207 & 1515 & 34,272 & 03/01/2012 – 06/27/2012 & Traffic speed\\
PEMS-BAY & 325 & 2369 & 52,116 & 01/01/2017 – 06/30/2017 & Traffic speed\\
PEMS03 & 358 & 547 & 26208 & 09/01/2018 – 11/30/2018 & Traffic flow\\
PEMS04 & 307 & 340 & 16992 & 01/01/2018 – 02/28/2018 & Traffic flow\\
PEMS07 & 883 & 866 & 28224 & 05/01/2017 – 08/06/2017 & Traffic flow\\
PEMS08 & 170 & 295 & 17856 & 07/01/2016 – 08/31/2016 & Traffic flow\\
\bottomrule
\end{tabular}
\label{tab:dataset}
\end{table}

\subsubsection{Evaluation Protocol.}
We adopt three standard metrics for evaluation: Mean Absolute Error (MAE), Root Mean Square Error (RMSE), and Mean Absolute Percentage Error (MAPE). For comprehensive assessment, we examine model performance across different prediction horizons (15-min, 30-min, and 60-min) and report both the horizon-specific and average results. Each experiment is repeated five times with different random seeds to ensure statistical reliability.

\subsubsection{Baselines.}
We evaluate six representative spatio-temporal prediction models as backbone networks to validate the effectiveness of the \model framework. Long Short-Term Memory (LSTM)~\cite{hochreiter1997long} controls the flow of information by introducing a gating mechanism to selectively keep and forget temporal data. Spatio-temporal Graph Convolution Network (STGCN)~\cite{yu2018spatio} combines graph convolution and 1D convolution to process spatio-temporal data. Graph WaveNet (GWN)~\cite{wu2019graph} utilizes adaptive adjacency matrices and dilation convolution to capture spatial and temporal correlations in traffic data. Adaptive Graph Convolutional Recurrent Network (AGCRN)~\cite{bai2020adaptive} infers dependencies between streaming time series through node adaptive parameter learning and data-adaptive graph generation modules. Decoupled Dynamic Spatio-Temporal Graph Neural Network (D2STGNN)~\cite{shao2022decoupled} is able to separate diffuse and intrinsic traffic information, thus enhancing dynamic graph learning. Finally, Spatio-Temporal Adaptive Embedding transformer (STAE)~\cite{liu2023spatio} encodes nodal, spatial, and temporal features through linear layers and multiple embedding layers respectively. 

\subsection{Model Settings of \model~(RQ1)}
All experiments are conducted using the PyTorch framework on a Linux server equipped with NVIDIA RTX A6000 GPUs. For model training, we employ Adam optimizer with an initial learning rate of 0.001 and weight decay of 0.0005. The evaluation metrics include MAE, RMSE, and MAPE, which are standard in traffic forecasting tasks. For comprehensive assessment, we reported the average performance across all 12 prediction horizons on four traffic flow datasets (PEMS03, PEMS04, PEMS07, and PEMS08).

We directly converted the original baselines to \model framework as backbones. The learning rate is set to be adjusted in step 10, and the ratio is 0.1. Various models are implemented concerning the benchmark LargeST~\cite{liu2024largest} and their official source code. For hyper-parameters in the framework, such as the number of NSPs is usually taken as 1, the number of node adaptive low-rank layers is usually taken as 4, and the maximum rank of the low-rank space is usually taken as 16, which is generally adjusted according to the original model. This setup ensures that we seamlessly integrate existing methods into our framework. Experiments show that our proposed framework supports multiple spatio-temporal prediction models, including the aforementioned baselines.

\subsection{Performance Comparisons (RQ2)}
To validate the effectiveness of our framework, we conducted comprehensive experiments across multiple models and datasets. Specifically, we analyze the performance improvements of different baseline models on the PEMS04 dataset (Table~\ref{tab:different_model}), and the generalization capability of our framework across multiple traffic datasets (Table~\ref{tab:different_dataset}). For statistical reliability, each experiment was repeated five times, with models enhanced by our framework denoted with a "+" suffix.

\begin{table*}[t]
\setlength{\tabcolsep}{0.2mm}{}
\centering
\caption{The improvement of different models in the PEMS04 dataset. Here, lower values indicate better performance. All six baselines have achieved significant improvements, denoted by $\Delta$. The subscripts indicate standard deviations.}
   \scalebox{0.6}{
    \begin{tabular}{c|ccc|ccc|ccc|ccc}
    \toprule
    \multirow{2}[4]{*}{Model} & \multicolumn{3}{c|}{15min} & \multicolumn{3}{c|}{30min} & \multicolumn{3}{c|}{60min} & \multicolumn{3}{c}{Average} \\
    \cmidrule{2-13}
    & MAE$~\downarrow$ & RMSE$~\downarrow$ & MAPE\%$~\downarrow$ & MAE$~\downarrow$ & RMSE$~\downarrow$ & MAPE\%$~\downarrow$ & MAE$~\downarrow$ & RMSE$~\downarrow$ & MAPE\%$~\downarrow$ & MAE$~\downarrow$ & RMSE$~\downarrow$ & MAPE\%$~\downarrow$ \\
    \midrule
    HA & 28.92$_{\pm1.25}$ & 42.69$_{\pm1.82}$ & 20.31$_{\pm0.89}$ & 33.73$_{\pm1.28}$ & 49.37$_{\pm1.85}$ & 24.01$_{\pm0.91}$ & 46.97$_{\pm1.31}$ & 67.43$_{\pm1.89}$ & 35.11$_{\pm0.92}$ & 38.03$_{\pm1.28}$ & 59.24$_{\pm1.85}$ & 27.88$_{\pm0.91}$ \\
    VAR & 21.94$_{\pm0.62}$ & 34.30$_{\pm1.02}$ & 16.42$_{\pm0.48}$ & 23.72$_{\pm0.71}$ & 36.58$_{\pm1.08}$ & 18.02$_{\pm0.52}$ & 26.76$_{\pm0.82}$ & 40.28$_{\pm1.23}$ & 20.94$_{\pm0.64}$ & 23.51$_{\pm0.72}$ & 36.39$_{\pm1.11}$ & 17.85$_{\pm0.55}$ \\
    SVR   & 22.52$_{\pm0.68}$ & 35.30$_{\pm1.12}$ & 14.71$_{\pm0.45}$ & 27.63$_{\pm0.78}$ & 42.23$_{\pm1.25}$ & 18.29$_{\pm0.49}$ & 37.86$_{\pm1.15}$ & 56.01$_{\pm1.70}$ & 26.72$_{\pm0.82}$ & 28.66$_{\pm0.87}$ & 44.59$_{\pm1.36}$ & 19.15$_{\pm0.59}$ \\
    \midrule
    LSTM  & 21.94$_{\pm0.59}$ & 33.37$_{\pm0.93}$ & 15.32$_{\pm0.40}$ & 25.83$_{\pm0.66}$ & 39.10$_{\pm1.04}$ & 20.35$_{\pm0.43}$ & 36.41$_{\pm0.82}$ & 50.73$_{\pm1.28}$ & 29.92$_{\pm0.56}$ & 27.14$_{\pm0.69}$ & 41.59$_{\pm1.08}$ & 18.20$_{\pm0.46}$ \\
    LSTM+ & 18.89$_{\pm0.58}$ & 29.96$_{\pm0.91}$ & 13.02$_{\pm0.40}$ & 21.31$_{\pm0.65}$ & 34.22$_{\pm1.04}$ & 13.96$_{\pm0.43}$ & 26.34$_{\pm0.80}$ & 41.30$_{\pm1.26}$ & 18.26$_{\pm0.56}$ & 22.18$_{\pm0.68}$ & 35.16$_{\pm1.07}$ & 15.08$_{\pm0.46}$ \\
    \rowcolor{green!20} \cellcolor{white!20} $\Delta$ & -3.05$_{\pm0.18}$ & -3.41$_{\pm0.20}$ & -2.30$_{\pm0.14}$ & -4.52$_{\pm0.27}$ & -4.88$_{\pm0.29}$ & -6.39$_{\pm0.38}$ & -10.07$_{\pm0.60}$ & -9.43$_{\pm0.57}$ & -11.66$_{\pm0.70}$ & -4.96$_{\pm0.30}$ & -6.43$_{\pm0.39}$ & -3.12$_{\pm0.19}$ \\
    \midrule
    STGCN & 19.45$_{\pm0.59}$ & 30.12$_{\pm0.92}$ & 14.21$_{\pm0.43}$ & 21.85$_{\pm0.62}$ & 34.43$_{\pm0.97}$ & 14.13$_{\pm0.44}$ & 26.97$_{\pm0.68}$ & 41.11$_{\pm1.06}$ & 16.84$_{\pm0.48}$ & 22.70$_{\pm0.63}$ & 35.55$_{\pm0.98}$ & 14.59$_{\pm0.45}$ \\
    STGCN+ & 19.12$_{\pm0.58}$ & 29.72$_{\pm0.91}$ & 13.89$_{\pm0.42}$ & 19.92$_{\pm0.61}$ & 31.63$_{\pm0.96}$ & 13.77$_{\pm0.42}$ & 22.07$_{\pm0.67}$ & 34.47$_{\pm1.05}$ & 15.42$_{\pm0.47}$ & 20.37$_{\pm0.62}$ & 31.94$_{\pm0.97}$ & 14.36$_{\pm0.44}$ \\
    \rowcolor{green!20} \cellcolor{white!20} $\Delta $ & -0.33$_{\pm0.02}$ & -0.40$_{\pm0.02}$ & -0.32$_{\pm0.02}$ & -1.93$_{\pm0.12}$ & -2.80$_{\pm0.17}$ & -0.36$_{\pm0.02}$ & -4.90$_{\pm0.29}$ & -6.64$_{\pm0.40}$ & -1.42$_{\pm0.09}$ & -2.33$_{\pm0.14}$ & -3.61$_{\pm0.22}$ & -0.23$_{\pm0.01}$ \\
    \midrule
    GWNet & 18.65$_{\pm0.57}$ & 29.24$_{\pm0.89}$ & 13.82$_{\pm0.42}$ & 19.57$_{\pm0.60}$ & 30.62$_{\pm0.92}$ & 13.28$_{\pm0.39}$ & 23.07$_{\pm0.70}$ & 35.35$_{\pm1.08}$ & 17.34$_{\pm0.53}$ & 25.45$_{\pm0.62}$ & 39.70$_{\pm0.97}$ & 17.29$_{\pm0.45}$ \\
    GWNet+ & 17.89$_{\pm0.55}$ & 28.52$_{\pm0.87}$ & 12.64$_{\pm0.39}$ & 18.88$_{\pm0.58}$ & 29.38$_{\pm0.89}$ & 13.06$_{\pm0.40}$ & 20.89$_{\pm0.64}$ & 32.96$_{\pm1.00}$ & 14.92$_{\pm0.46}$ & 19.22$_{\pm0.59}$ & 30.62$_{\pm0.93}$ & 13.54$_{\pm0.41}$ \\
    \rowcolor{green!20} \cellcolor{white!20} $\Delta$ & -0.76$_{\pm0.05}$ & -0.72$_{\pm0.04}$ & -1.18$_{\pm0.07}$ & -0.69$_{\pm0.04}$ & -1.24$_{\pm0.07}$ & -0.22$_{\pm0.01}$ & -2.18$_{\pm0.13}$ & -2.39$_{\pm0.14}$ & -2.42$_{\pm0.15}$ & -6.23$_{\pm0.37}$ & -9.08$_{\pm0.54}$ & -3.75$_{\pm0.23}$ \\
    \midrule
    AGCRN & 18.12$_{\pm0.55}$ & 29.45$_{\pm0.90}$ & 12.85$_{\pm0.39}$ & 18.77$_{\pm0.57}$ & 30.08$_{\pm0.92}$ & 12.97$_{\pm0.40}$ & 20.41$_{\pm0.62}$ & 32.87$_{\pm1.00}$ & 14.38$_{\pm0.44}$ & 19.83$_{\pm0.58}$ & 32.26$_{\pm0.94}$ & 13.40$_{\pm0.41}$ \\
    AGCRN+ & 17.83$_{\pm0.54}$ & 29.16$_{\pm0.89}$ & 12.55$_{\pm0.38}$ & 18.63$_{\pm0.57}$ & 29.99$_{\pm0.91}$ & 12.82$_{\pm0.39}$ & 19.97$_{\pm0.61}$ & 32.37$_{\pm0.99}$ & 13.78$_{\pm0.42}$ & 18.81$_{\pm0.57}$ & 30.51$_{\pm0.93}$ & 13.05$_{\pm0.40}$ \\
    \rowcolor{green!20} \cellcolor{white!20} $\Delta$ & -0.29$_{\pm0.02}$ & -0.29$_{\pm0.02}$ & -0.30$_{\pm0.02}$ & -0.14$_{\pm0.01}$ & -0.09$_{\pm0.01}$ & -0.15$_{\pm0.01}$ & -0.44$_{\pm0.03}$ & -0.50$_{\pm0.03}$ & -0.60$_{\pm0.04}$ & -1.02$_{\pm0.06}$ & -1.75$_{\pm0.11}$ & -0.35$_{\pm0.02}$ \\
    \midrule
    STAE  & 17.95$_{\pm0.55}$ & 29.12$_{\pm0.89}$ & 12.65$_{\pm0.39}$ & 18.92$_{\pm0.58}$ & 30.09$_{\pm0.92}$ & 13.35$_{\pm0.41}$ & 21.06$_{\pm0.64}$ & 33.37$_{\pm1.02}$ & 15.55$_{\pm0.47}$ & 19.31$_{\pm0.59}$ & 30.86$_{\pm0.94}$ & 13.85$_{\pm0.42}$ \\
    STAE+ & 17.65$_{\pm0.54}$ & 28.73$_{\pm0.88}$ & 12.45$_{\pm0.38}$ & 18.62$_{\pm0.57}$ & 29.55$_{\pm0.90}$ & 13.29$_{\pm0.41}$ & 20.40$_{\pm0.62}$ & 32.38$_{\pm0.99}$ & 15.00$_{\pm0.46}$ & 18.89$_{\pm0.58}$ & 30.22$_{\pm0.92}$ & 13.58$_{\pm0.41}$ \\
    \rowcolor{green!20} \cellcolor{white!20} $\Delta$ & -0.30$_{\pm0.02}$ & -0.39$_{\pm0.02}$ & -0.20$_{\pm0.01}$ & -0.30$_{\pm0.02}$ & -0.54$_{\pm0.03}$ & -0.06$_{\pm0.01}$ & -0.66$_{\pm0.04}$ & -0.99$_{\pm0.06}$ & -0.55$_{\pm0.03}$ & -0.42$_{\pm0.03}$ & -0.64$_{\pm0.04}$ & -0.27$_{\pm0.02}$ \\
    \midrule
    D2STGNN & 18.95$_{\pm0.58}$ & 29.85$_{\pm0.91}$ & 14.82$_{\pm0.45}$ & 19.96$_{\pm0.61}$ & 31.34$_{\pm0.95}$ & 15.52$_{\pm0.47}$ & 23.34$_{\pm0.71}$ & 35.89$_{\pm1.09}$ & 17.39$_{\pm0.53}$ & 20.75$_{\pm0.63}$ & 32.36$_{\pm0.99}$ & 15.91$_{\pm0.49}$ \\
    D2STGNN+ & 18.25$_{\pm0.56}$ & 28.92$_{\pm0.88}$ & 14.12$_{\pm0.43}$ & 19.21$_{\pm0.59}$ & 30.50$_{\pm0.93}$ & 13.46$_{\pm0.41}$ & 21.73$_{\pm0.66}$ & 33.73$_{\pm1.03}$ & 17.00$_{\pm0.52}$ & 19.73$_{\pm0.60}$ & 31.05$_{\pm0.95}$ & 14.86$_{\pm0.45}$ \\
    \rowcolor{green!20} \cellcolor{white!20} $\Delta$ & -0.70$_{\pm0.04}$ & -0.93$_{\pm0.06}$ & -0.70$_{\pm0.04}$ & -0.75$_{\pm0.05}$ & -0.84$_{\pm0.05}$ & -2.06$_{\pm0.12}$ & -1.61$_{\pm0.10}$ & -2.16$_{\pm0.13}$ & -0.39$_{\pm0.02}$ & -1.02$_{\pm0.06}$ & -1.31$_{\pm0.08}$ & -1.05$_{\pm0.06}$ \\
    \bottomrule
    \end{tabular}
    }
  \label{tab:different_model}
\end{table*}
\begin{table}[h!]
  \centering
  \caption{Performance Improvements of one of the backbone STGNNs on Multiple Traffic Datasets. We use STGCN in the table as an example to illustrate the significant enhancement of our method from the perspective of the dataset.}
  \scalebox{0.62}{
    \begin{tabular}{c|ccc|ccc|ccc|ccc}
    \toprule
    \multirow{2}{*}{Dataset} & \multicolumn{3}{c|}{15min} & \multicolumn{3}{c|}{30min} & \multicolumn{3}{c|}{60min} & \multicolumn{3}{c}{Average} \\
    \cmidrule{2-13}
     & MAE & RMSE & MAPE\% & MAE & RMSE & MAPE\% & MAE & RMSE & MAPE\% & MAE & RMSE & MAPE\% \\
    \midrule
    PEMS04 & -0.33$_{\pm0.02}$ & -0.40$_{\pm0.03}$ & -0.07$_{\pm0.01}$ & -1.45$_{\pm0.09}$ & -2.50$_{\pm0.15}$ & -0.10$_{\pm0.01}$ & -4.94$_{\pm0.30}$ & -6.61$_{\pm0.40}$ & -1.28$_{\pm0.08}$ & -2.24$_{\pm0.14}$ & -3.17$_{\pm0.19}$ & -0.48$_{\pm0.03}$ \\
    \midrule
    PEMS08 & -0.20$_{\pm0.01}$ & -1.09$_{\pm0.07}$ & -0.57$_{\pm0.03}$ & -1.37$_{\pm0.08}$ & -1.79$_{\pm0.11}$ & -0.30$_{\pm0.02}$ & -7.63$_{\pm0.46}$ & -5.86$_{\pm0.35}$ & -1.28$_{\pm0.08}$ & -3.07$_{\pm0.18}$ & -2.91$_{\pm0.17}$ & -0.72$_{\pm0.04}$ \\
    \midrule
    PEMS03 & -0.31$_{\pm0.02}$ & -0.51$_{\pm0.03}$ & -0.72$_{\pm0.04}$ & -0.22$_{\pm0.01}$ & -0.23$_{\pm0.01}$ & -0.60$_{\pm0.04}$ & -0.45$_{\pm0.03}$ & -0.35$_{\pm0.02}$ & -1.25$_{\pm0.08}$ & -0.33$_{\pm0.02}$ & -0.36$_{\pm0.02}$ & -0.86$_{\pm0.05}$ \\
    \midrule
    PEMS07 & -0.26$_{\pm0.02}$ & -0.29$_{\pm0.02}$ & -0.16$_{\pm0.01}$ & -0.38$_{\pm0.02}$ & -0.35$_{\pm0.02}$ & -0.31$_{\pm0.02}$ & -0.59$_{\pm0.04}$ & -0.59$_{\pm0.04}$ & -0.32$_{\pm0.02}$ & -0.41$_{\pm0.03}$ & -0.41$_{\pm0.03}$ & -0.26$_{\pm0.02}$ \\
    \midrule
    METR-LA & -0.10$_{\pm0.01}$ & -0.36$_{\pm0.02}$ & -0.03$_{\pm0.00}$ & -0.33$_{\pm0.02}$ & -0.83$_{\pm0.05}$ & -0.80$_{\pm0.05}$ & -1.05$_{\pm0.06}$ & -1.97$_{\pm0.12}$ & -2.30$_{\pm0.14}$ & -0.49$_{\pm0.03}$ & -1.05$_{\pm0.06}$ & -1.04$_{\pm0.06}$ \\
    \midrule
    PEMSBAY & -0.04$_{\pm0.00}$ & -0.06$_{\pm0.00}$ & -0.11$_{\pm0.01}$ & -0.10$_{\pm0.01}$ & -0.45$_{\pm0.03}$ & -0.23$_{\pm0.01}$ & -0.50$_{\pm0.03}$ & -1.14$_{\pm0.07}$ & -1.02$_{\pm0.06}$ & -0.21$_{\pm0.01}$ & -0.55$_{\pm0.03}$ & -0.45$_{\pm0.03}$ \\
    \bottomrule
    \end{tabular}%
  }
  \label{tab:different_dataset}
\end{table}

As shown in Table~\ref{tab:different_model}, \model demonstrates consistent performance improvements across various baseline models, ranging from traditional models to state-of-the-art approaches. For traditional methods, LSTM achieves remarkable improvements with MAE reductions of 3.05, 4.52, and 10.07 at 15-min, 30-min, and 60-min horizons, respectively. The enhancement is particularly evident in long-term predictions, where STGCN shows MAE reductions of 0.33, 1.93, and 4.90 across different horizons. Even sophisticated models like D2STGNN and AGCRN benefit from our framework, with D2STGNN achieving consistent MAE reductions of 0.70, 0.75, and 1.61, while AGCRN shows stable improvements with an average MAE reduction of 1.02. These results validate the effectiveness of our framework in enhancing STGNNs through node-level low-rank adaptations.

The results in Table~\ref{tab:different_dataset} demonstrate our framework's generalization capability across diverse datasets. Using STGCN as an example, we observe substantial improvements on all datasets. On PEMS04, MAE reductions of 0.33, 1.45, and 4.94 are achieved for 15-min, 30-min, and 60-min horizons respectively, with an average reduction of 2.24. PEMS08 shows even stronger improvements with MAE reductions up to 7.63 for 60-min predictions. The effectiveness varies systematically across different prediction horizons, with longer-term predictions (60-min) consistently exhibiting more significant improvements. On datasets with distinct spatial structures (PEMS03, PEMS07, METR-LA, and PEMS-BAY), the framework maintains robust performance gains. These improvements across datasets with varying node counts confirm the framework's versatile adaptability to different types of spatio-temporal data.

The superior performance can be attributed to two key aspects. First, our node-adaptive approach provides a precisely calibrated parameter space for fine-tuning predictions while maintaining computational efficiency through low-rank matrix factorization. Second, the additional parameters from low-rank matrices effectively capture complex spatio-temporal dependencies, including regional characteristics, temporal dynamics, and node interactions, all achieved within a compressed parameter space. This synergistic combination of adaptive capacity and computational efficiency enables our framework to enhance various baseline models consistently across diverse prediction scenarios.

\begin{figure}[!b]
  \centering
  \subfloat[Processing time.]{
    \includegraphics[width=0.45\linewidth]{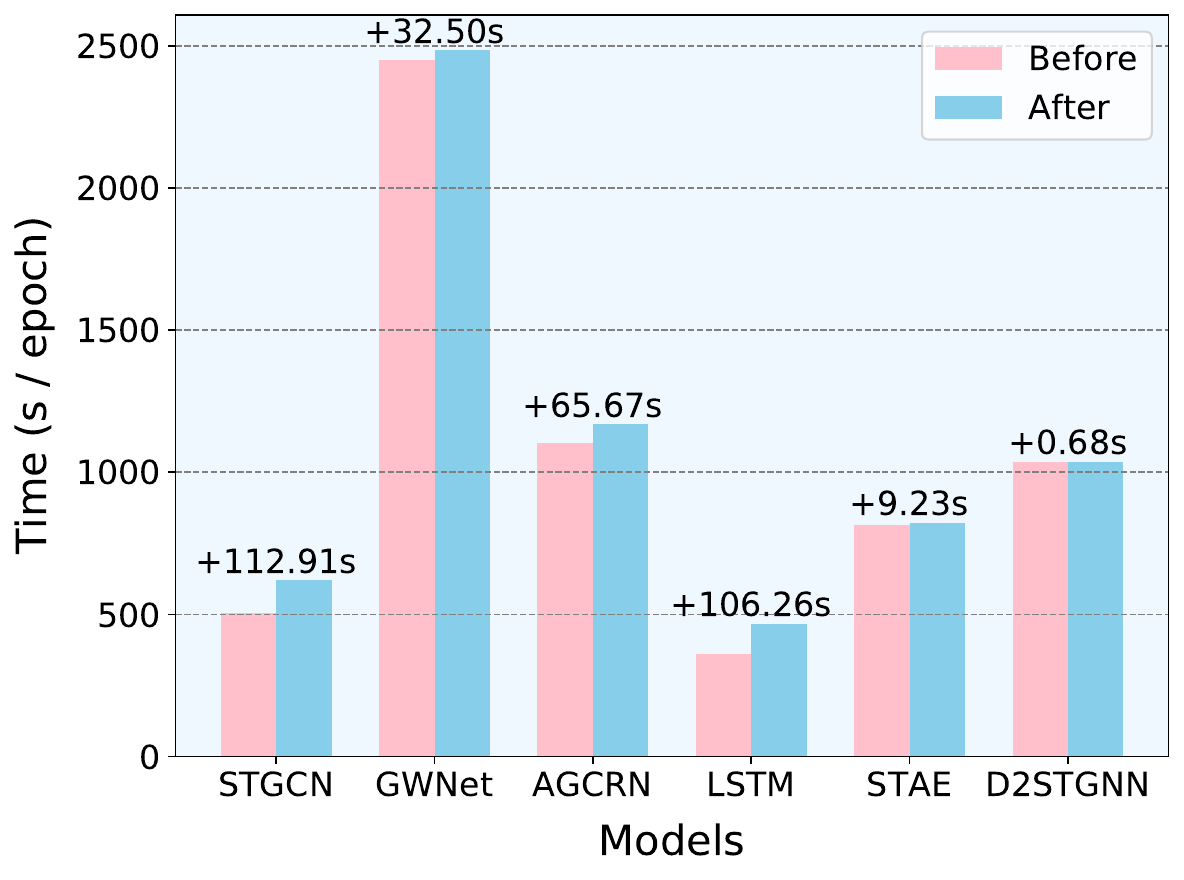}
  }
  \subfloat[Training parameter.]{
    \includegraphics[width=0.45\linewidth]{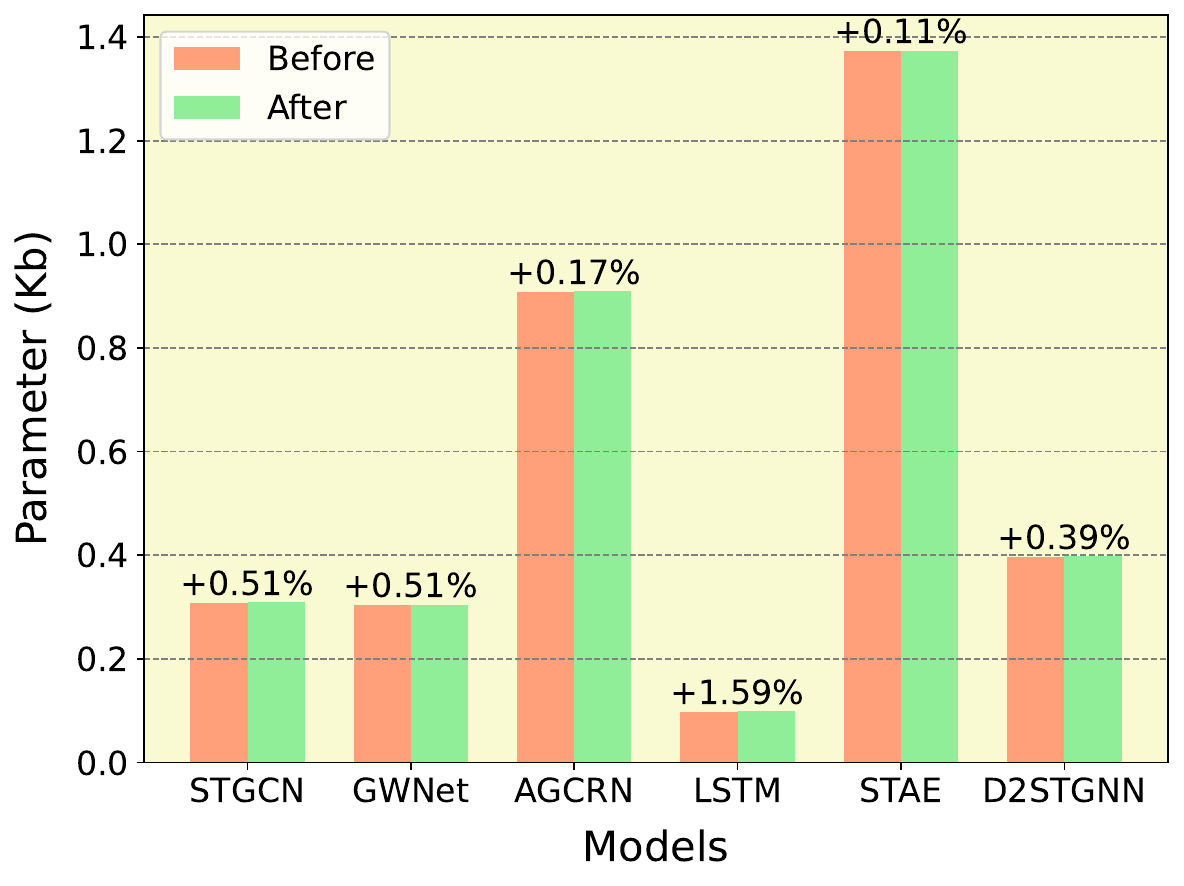}
  }
  \caption{Efficiency study comparing time and parameter cost of ST-LoRA.}
  \label{fig:efficiency_compare}
\end{figure}

\subsection{Efficiency and scalability Studies (RQ3)}
\subsubsection{Time Efficiency.} In Figure~\ref{fig:efficiency_compare}(a), we analyze the computational overhead when integrating our framework with existing models. The results demonstrate that ST-LoRA introduces minimal additional training time while delivering substantial performance gains. With 16 NALLs and 4 MLRFs to enhance, the time increase remains remarkably efficient, adding a mere 0.68 seconds for D2STGNN. This efficiency stems from our low-rank adaptation strategy, which maintains stable training times even with multiple node-specific predictors.

\subsubsection{Framework Scalability.} We evaluate parameter efficiency across six baseline models using consistent node adaptive low-rank layer configurations. As shown in Figure~\ref{fig:efficiency_compare}(b), our approach achieves remarkable parameter efficiency, requiring less than 1\% additional trainable parameters for most models while delivering 4.2\% to 7.3\% improvements in average RMSE. Notably, the smallest LSTM model maintains overhead below 2\% while achieving an impressive 15\% reduction in Average RMSE. This demonstrates ST-LoRA's ability to effectively capture and adapt to node heterogeneity with minimal computational overhead.

\begin{figure}[!t]
    \centering
    \includegraphics[width=.95\linewidth]{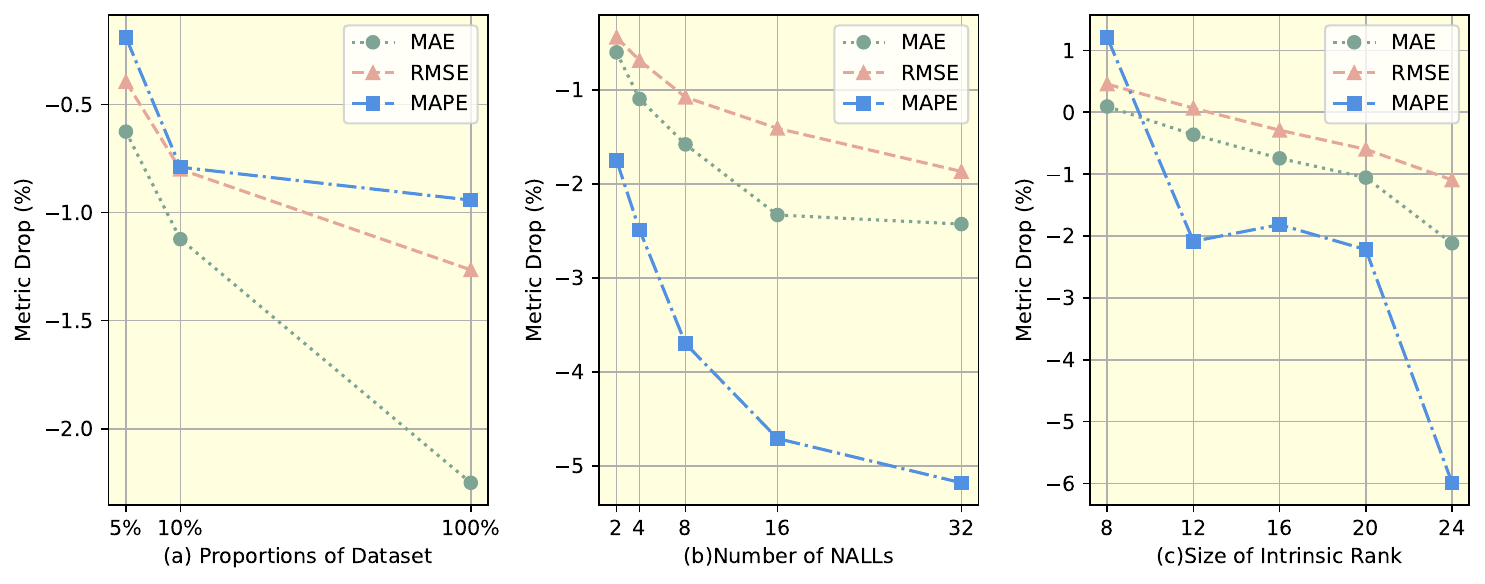}
    \caption{Parameter sensitivity analysis examining the impact of varying dimensions, placements, and quantities of adaptive components on model performance.}
    \label{fig:para_study}
\end{figure}
\subsection{Parameter Sensitivity Analysis (RQ4)}
In our experiments, we fixed the number of layers at 4 and the node embedding dimension at 12, which are the two most critical parameters. The node embedding dimension represents the rank of the low-rank matrix required for the additional parameters of the nodes, which tends to increase as more feature information is included in the data. After stacking multiple layers of NALL, the fine-tuning effect of these additional parameter spaces is amplified. It is important to design these two hyper-parameters in a balanced way because a larger parameter space does not necessarily mean it is easier to learn. We again use the STGCN model and the PEMS04 dataset as an example to explore the relationship between the size of this dataset as well as the two hyper-parameters and the lifting effect, and the results are shown in Figure~\ref{fig:para_study}.

\section{Conclusion}
In this paper, we focus on the challenges of node-level heterogeneity and overparameterization in spatio-temporal modeling. We introduce \model, a novel framework featuring Node-Adaptive Low-rank Layers and Node-Specific Predictors that efficiently customize parameters while maintaining computational efficiency. Our proposed approach enhances existing models by effectively capturing heterogeneous features and distributional changes within independent nodes. The improvements demonstrate that efficient low-rank adaptation can significantly enhance forecasting in domains with heterogeneous node behaviors.

\section{Limitation and Future work}
Despite the promising improvements brought by \model, the development and application of Low-rank adaptation techniques have so far been mostly concentrated within large models. When applied to specialized spatio-temporal models, adapting our method often requires substantial manual intervention, which collectively hinder the straightforward and scalable adoption of LoRA across heterogeneous spatio-temporal forecasting models.
To overcome these limitations, we plan to continuously update and maintain a comprehensive benchmark that includes a wide variety of spatio-temporal models and datasets. Specifically, our future work will focus on: 1) developing a unified framework that enables flexible and automated adaptation of LoRA structures to different baseline models, supporting multiple LoRA variants for systematic and reproducible comparison; 
and 2) conduct extensive experiments across diverse real-world spatio-temporal scenarios, thereby providing the research community with a fair and comprehensive baseline for future studies.

\section*{Acknowledgment}
This work is supported by the National Natural Science Foundation of China (No. 62402414), the Guangdong Basic and Applied Basic Research Foundation (No. 2025A1515011994), and a grant from State Key Laboratory of Resources and Environmental Information System. This study is also supported by the Guangzhou Municipal Science and Technology Project (No. 2023A03J0011), the Guangzhou Industrial Information and Intelligent Key Laboratory Project (No. 2024A03J0628), and Guangdong Provincial Key Lab of Integrated Communication, Sensing and Computation for Ubiquitous Internet of Things (No. 2023B1212010007).

\bibliographystyle{plain}

\begin{thebibliography}{10}

\bibitem{agiza2024mtlora}
Ahmed Agiza, Marina Neseem, and Sherief Reda.
\newblock Mtlora: Low-rank adaptation approach for efficient multi-task learning.
\newblock In {\em Proceedings of the IEEE/CVF conference on computer vision and pattern recognition}, pages 16196--16205, 2024.

\bibitem{bai2020adaptive}
Lei Bai, Lina Yao, Can Li, Xianzhi Wang, and Can Wang.
\newblock Adaptive graph convolutional recurrent network for traffic forecasting.
\newblock {\em Advances in neural information processing systems}, 33:17804--17815, 2020.

\bibitem{chen2001freeway}
Chao Chen, Karl Petty, Alexander Skabardonis, Pravin Varaiya, and Zhanfeng Jia.
\newblock Freeway performance measurement system: mining loop detector data.
\newblock {\em Transportation Research Record}, 1748(1):96--102, 2001.

\bibitem{chung2014empirical}
Junyoung Chung, Caglar Gulcehre, KyungHyun Cho, and Yoshua Bengio.
\newblock Empirical evaluation of gated recurrent neural networks on sequence modeling, 2014.

\bibitem{dong2024heterogeneity}
Zheng Dong, Renhe Jiang, Haotian Gao, Hangchen Liu, Jinliang Deng, Qingsong Wen, and Xuan Song.
\newblock Heterogeneity-informed meta-parameter learning for spatiotemporal time series forecasting.
\newblock In {\em Proceedings of the 30th ACM SIGKDD Conference on Knowledge Discovery and Data Mining}, 2024.

\bibitem{du2018sensable}
Rong Du, Paolo Santi, Ming Xiao, Athanasios~V Vasilakos, and Carlo Fischione.
\newblock The sensable city: A survey on the deployment and management for smart city monitoring.
\newblock {\em IEEE Communications Surveys \& Tutorials}, 21(2):1533--1560, 2018.

\bibitem{fang2021spatial}
Zheng Fang, Qingqing Long, Guojie Song, and Kunqing Xie.
\newblock Spatial-temporal graph ode networks for traffic flow forecasting.
\newblock In {\em Proceedings of the 27th ACM SIGKDD conference on knowledge discovery \& data mining}, pages 364--373, 2021.

\bibitem{gao2023spatial}
Haotian Gao, Renhe Jiang, Zheng Dong, Jinliang Deng, Yuxin Ma, and Xuan Song.
\newblock Spatial-temporal-decoupled masked pre-training for spatiotemporal forecasting.
\newblock {\em arXiv preprint arXiv:2312.00516}, 2023.

\bibitem{gu2018recent}
Jiuxiang Gu, Zhenhua Wang, Jason Kuen, Lianyang Ma, Amir Shahroudy, Bing Shuai, Ting Liu, Xingxing Wang, Gang Wang, Jianfei Cai, et~al.
\newblock Recent advances in convolutional neural networks.
\newblock {\em Pattern recognition}, 77:354--377, 2018.

\bibitem{hayou2024lora+}
Soufiane Hayou, Nikhil Ghosh, and Bin Yu.
\newblock Lora+: Efficient low rank adaptation of large models.
\newblock {\em arXiv preprint arXiv:2402.12354}, 2024.

\bibitem{hochreiter1997long}
Sepp Hochreiter and J{\"u}rgen Schmidhuber.
\newblock Long short-term memory.
\newblock {\em Neural computation}, 9(8):1735--1780, 1997.

\bibitem{hu2021lora}
Edward~J Hu, Yelong Shen, Phillip Wallis, Zeyuan Allen-Zhu, Yuanzhi Li, Shean Wang, Lu~Wang, and Weizhu Chen.
\newblock Lora: Low-rank adaptation of large language models, 2021.

\bibitem{huang2014deep}
Wenhao Huang, Guojie Song, Haikun Hong, and Kunqing Xie.
\newblock Deep architecture for traffic flow prediction: deep belief networks with multitask learning.
\newblock {\em IEEE Transactions on Intelligent Transportation Systems}, 15(5):2191--2201, 2014.

\bibitem{jhuo2012robust}
I-Hong Jhuo, Dong Liu, DT~Lee, and Shih-Fu Chang.
\newblock Robust visual domain adaptation with low-rank reconstruction.
\newblock In {\em 2012 IEEE conference on computer vision and pattern recognition}, pages 2168--2175. IEEE, 2012.

\bibitem{jiang2023spatio}
Renhe Jiang, Zhaonan Wang, Jiawei Yong, Puneet Jeph, Quanjun Chen, Yasumasa Kobayashi, Xuan Song, Shintaro Fukushima, and Toyotaro Suzumura.
\newblock Spatio-temporal meta-graph learning for traffic forecasting.
\newblock In {\em Proceedings of the AAAI conference on artificial intelligence}, volume~37, pages 8078--8086, 2023.

\bibitem{jin2023spatio}
Guangyin Jin, Yuxuan Liang, Yuchen Fang, Zezhi Shao, Jincai Huang, Junbo Zhang, and Yu~Zheng.
\newblock Spatio-temporal graph neural networks for predictive learning in urban computing: A survey.
\newblock {\em IEEE Transactions on Knowledge and Data Engineering}, 2023.

\bibitem{jin2024position}
Ming Jin, Yifan Zhang, Wei Chen, Kexin Zhang, Yuxuan Liang, Bin Yang, Jindong Wang, Shirui Pan, and Qingsong Wen.
\newblock Position: What can large language models tell us about time series analysis.
\newblock In {\em Forty-first International Conference on Machine Learning}, 2024.

\bibitem{karimi2021compacter}
Rabeeh Karimi~Mahabadi, James Henderson, and Sebastian Ruder.
\newblock Compacter: Efficient low-rank hypercomplex adapter layers.
\newblock {\em Advances in Neural Information Processing Systems}, 34:1022--1035, 2021.

\bibitem{kipf2016semi}
Thomas~N Kipf and Max Welling.
\newblock Semi-supervised classification with graph convolutional networks, 2016.

\bibitem{li2017diffusion}
Yaguang Li, Rose Yu, Cyrus Shahabi, and Yan Liu.
\newblock Diffusion convolutional recurrent neural network: Data-driven traffic forecasting, 2017.

\bibitem{li2018algorithmic}
Yuanzhi Li, Tengyu Ma, and Hongyang Zhang.
\newblock Algorithmic regularization in over-parameterized matrix sensing and neural networks with quadratic activations.
\newblock In {\em Conference On Learning Theory}, pages 2--47. PMLR, 2018.

\bibitem{li2022spatial}
Zhonghang Li, Chao Huang, Lianghao Xia, Yong Xu, and Jian Pei.
\newblock Spatial-temporal hypergraph self-supervised learning for crime prediction.
\newblock In {\em 2022 IEEE 38th international conference on data engineering (ICDE)}, pages 2984--2996. IEEE, 2022.

\bibitem{liang2018geoman}
Yuxuan Liang, Songyu Ke, Junbo Zhang, Xiuwen Yi, and Yu~Zheng.
\newblock Geoman: Multi-level attention networks for geo-sensory time series prediction.
\newblock In {\em IJCAI}, volume 2018, pages 3428--3434, 2018.

\bibitem{liang2021revisiting}
Yuxuan Liang, Kun Ouyang, Yiwei Wang, Ye~Liu, Junbo Zhang, Yu~Zheng, and David~S Rosenblum.
\newblock Revisiting convolutional neural networks for citywide crowd flow analytics.
\newblock In {\em Machine Learning and Knowledge Discovery in Databases: European Conference, ECML PKDD 2020, Ghent, Belgium, September 14--18, 2020, Proceedings, Part I}, pages 578--594. Springer, 2021.

\bibitem{liang2023airformer}
Yuxuan Liang, Yutong Xia, Songyu Ke, Yiwei Wang, Qingsong Wen, Junbo Zhang, Yu~Zheng, and Roger Zimmermann.
\newblock Airformer: Predicting nationwide air quality in china with transformers.
\newblock In {\em Proceedings of the AAAI Conference on Artificial Intelligence}, pages 14329--14337, 2023.

\bibitem{lippi2013short}
Marco Lippi, Matteo Bertini, and Paolo Frasconi.
\newblock Short-term traffic flow forecasting: An experimental comparison of time-series analysis and supervised learning.
\newblock {\em IEEE Transactions on Intelligent Transportation Systems}, 14(2):871--882, 2013.

\bibitem{liu2023spatio}
Hangchen Liu, Zheng Dong, Renhe Jiang, Jiewen Deng, Jinliang Deng, Quanjun Chen, and Xuan Song.
\newblock Spatio-temporal adaptive embedding makes vanilla transformer sota for traffic forecasting.
\newblock In {\em Proceedings of the 32nd ACM international conference on information and knowledge management}, pages 4125--4129, 2023.

\bibitem{liu2024largest}
Xu~Liu, Yutong Xia, Yuxuan Liang, Junfeng Hu, Yiwei Wang, Lei Bai, Chao Huang, Zhenguang Liu, Bryan Hooi, and Roger Zimmermann.
\newblock Largest: A benchmark dataset for large-scale traffic forecasting.
\newblock {\em Advances in Neural Information Processing Systems}, 36, 2024.

\bibitem{lv2014traffic}
Yisheng Lv, Yanjie Duan, Wenwen Kang, Zhengxi Li, and Fei-Yue Wang.
\newblock Traffic flow prediction with big data: A deep learning approach.
\newblock {\em IEEE Transactions on Intelligent Transportation Systems}, 16(2):865--873, 2014.

\bibitem{ma2024spatio}
Ying Ma, Haijie Lou, Ming Yan, Fanghui Sun, and Guoqi Li.
\newblock Spatio-temporal fusion graph convolutional network for traffic flow forecasting.
\newblock {\em Information Fusion}, 104:102196, 2024.

\bibitem{pan2022st}
Junting Pan, Ziyi Lin, Xiatian Zhu, Jing Shao, and Hongsheng Li.
\newblock St-adapter: Parameter-efficient image-to-video transfer learning.
\newblock {\em Advances in Neural Information Processing Systems}, 35:26462--26477, 2022.

\bibitem{pan2019matrix}
Zheyi Pan, Zhaoyuan Wang, Weifeng Wang, Yong Yu, Junbo Zhang, and Yu~Zheng.
\newblock Matrix factorization for spatio-temporal neural networks with applications to urban flow prediction.
\newblock In {\em Proceedings of the 28th ACM international conference on information and knowledge management}, pages 2683--2691, 2019.

\bibitem{sainath2013low}
Tara~N Sainath, Brian Kingsbury, Vikas Sindhwani, Ebru Arisoy, and Bhuvana Ramabhadran.
\newblock Low-rank matrix factorization for deep neural network training with high-dimensional output targets.
\newblock In {\em 2013 IEEE international conference on acoustics, speech and signal processing}, pages 6655--6659. IEEE, 2013.

\bibitem{shao2022pre}
Zezhi Shao, Zhao Zhang, Fei Wang, and Yongjun Xu.
\newblock Pre-training enhanced spatial-temporal graph neural network for multivariate time series forecasting.
\newblock In {\em Proceedings of the 28th ACM SIGKDD Conference on Knowledge Discovery and Data Mining}, pages 1567--1577, 2022.

\bibitem{shao2022decoupled}
Zezhi Shao, Zhao Zhang, Wei Wei, Fei Wang, Yongjun Xu, Xin Cao, and Christian~S Jensen.
\newblock Decoupled dynamic spatial-temporal graph neural network for traffic forecasting.
\newblock {\em Proceedings of the VLDB Endowment}, 15(11):2733--2746, 2022.

\bibitem{smith1997traffic}
Brian~L Smith and Michael~J Demetsky.
\newblock Traffic flow forecasting: comparison of modeling approaches.
\newblock {\em Journal of transportation engineering}, 123(4):261--266, 1997.

\bibitem{valipour2022dylora}
Mojtaba Valipour, Mehdi Rezagholizadeh, Ivan Kobyzev, and Ali Ghodsi.
\newblock Dylora: Parameter efficient tuning of pre-trained models using dynamic search-free low-rank adaptation.
\newblock {\em arXiv preprint arXiv:2210.07558}, 2022.

\bibitem{van2012short}
JWC Van~Lint and CPIJ Van~Hinsbergen.
\newblock Short-term traffic and travel time prediction models.
\newblock {\em Artificial Intelligence Applications to Critical Transportation Issues}, 22(1):22--41, 2012.

\bibitem{vaswani2017attention}
Ashish Vaswani, Noam Shazeer, Niki Parmar, Jakob Uszkoreit, Llion Jones, Aidan~N Gomez, {\L}ukasz Kaiser, and Illia Polosukhin.
\newblock Attention is all you need.
\newblock {\em Advances in neural information processing systems}, 30, 2017.

\bibitem{wang2022hierarchical}
Hanqiu Wang, Rongqing Zhang, Xiang Cheng, and Liuqing Yang.
\newblock Hierarchical traffic flow prediction based on spatial-temporal graph convolutional network.
\newblock {\em IEEE Transactions on Intelligent Transportation Systems}, 23(9):16137--16147, 2022.

\bibitem{wang2020deep}
Senzhang Wang, Jiannong Cao, and S~Yu Philip.
\newblock Deep learning for spatio-temporal data mining: A survey.
\newblock {\em IEEE transactions on knowledge and data engineering}, 34(8):3681--3700, 2020.

\bibitem{wang2020traffic}
Xiaoyang Wang, Yao Ma, Yiqi Wang, Wei Jin, Xin Wang, Jiliang Tang, Caiyan Jia, and Jian Yu.
\newblock Traffic flow prediction via spatial temporal graph neural network.
\newblock In {\em Proceedings of the web conference 2020}, pages 1082--1092, 2020.

\bibitem{wu2021autoformer}
Haixu Wu, Jiehui Xu, Jianmin Wang, and Mingsheng Long.
\newblock Autoformer: Decomposition transformers with auto-correlation for long-term series forecasting.
\newblock {\em Advances in neural information processing systems}, 34:22419--22430, 2021.

\bibitem{wu2020comprehensive}
Zonghan Wu, Shirui Pan, Fengwen Chen, Guodong Long, Chengqi Zhang, and S~Yu Philip.
\newblock A comprehensive survey on graph neural networks.
\newblock {\em IEEE transactions on neural networks and learning systems}, 32(1):4--24, 2020.

\bibitem{wu2019graph}
Zonghan Wu, Shirui Pan, Guodong Long, Jing Jiang, and Chengqi Zhang.
\newblock Graph wavenet for deep spatial-temporal graph modeling.
\newblock In {\em Proceedings of the 28th International Joint Conference on Artificial Intelligence}, pages 1907--1913, 2019.

\bibitem{yan2024urbanclip}
Yibo Yan, Haomin Wen, Siru Zhong, Wei Chen, Haodong Chen, Qingsong Wen, Roger Zimmermann, and Yuxuan Liang.
\newblock Urbanclip: Learning text-enhanced urban region profiling with contrastive language-image pretraining from the web.
\newblock In {\em Proceedings of the ACM on Web Conference 2024}, pages 4006--4017, 2024.

\bibitem{yao2018deep}
Huaxiu Yao, Fei Wu, Jintao Ke, Xianfeng Tang, Yitian Jia, Siyu Lu, Pinghua Gong, Jieping Ye, and Zhenhui Li.
\newblock Deep multi-view spatial-temporal network for taxi demand prediction.
\newblock In {\em Proceedings of the AAAI conference on artificial intelligence}, 2018.

\bibitem{yu2018spatio}
Bing Yu, Haoteng Yin, and Zhanxing Zhu.
\newblock Spatio-temporal graph convolutional networks: a deep learning framework for traffic forecasting.
\newblock In {\em Proceedings of the 27th International Joint Conference on Artificial Intelligence}, pages 3634--3640, 2018.

\bibitem{yu2019review}
Yong Yu, Xiaosheng Si, Changhua Hu, and Jianxun Zhang.
\newblock A review of recurrent neural networks: Lstm cells and network architectures.
\newblock {\em Neural computation}, 31(7):1235--1270, 2019.

\bibitem{zhang2018combining}
Da~Zhang and Mansur~R Kabuka.
\newblock Combining weather condition data to predict traffic flow: a gru-based deep learning approach.
\newblock {\em IET Intelligent Transport Systems}, 12(7):578--585, 2018.

\bibitem{zhang2011data}
Junping Zhang, Fei-Yue Wang, Kunfeng Wang, Wei-Hua Lin, Xin Xu, and Cheng Chen.
\newblock Data-driven intelligent transportation systems: A survey.
\newblock {\em IEEE Transactions on Intelligent Transportation Systems}, 12(4):1624--1639, 2011.

\bibitem{zhang2020spatio}
Qi~Zhang, Jianlong Chang, Gaofeng Meng, Shiming Xiang, and Chunhong Pan.
\newblock Spatio-temporal graph structure learning for traffic forecasting.
\newblock In {\em Proceedings of the AAAI conference on artificial intelligence}, volume~34, pages 1177--1185, 2020.

\bibitem{zhou2022fedformer}
Tian Zhou, Ziqing Ma, Qingsong Wen, Xue Wang, Liang Sun, and Rong Jin.
\newblock Fedformer: Frequency enhanced decomposed transformer for long-term series forecasting.
\newblock In {\em International conference on machine learning}, pages 27268--27286. PMLR, 2022.

\bibitem{zhou2024one}
Tian Zhou, Peisong Niu, Liang Sun, Rong Jin, et~al.
\newblock One fits all: Power general time series analysis by pretrained lm.
\newblock {\em Advances in neural information processing systems}, 36, 2024.

\bibitem{zivot2006vector}
Eric Zivot and Jiahui Wang.
\newblock Vector autoregressive models for multivariate time series.
\newblock {\em Modeling financial time series with S-PLUS{\textregistered}}, pages 385--429, 2006.

\end{thebibliography}

\appendix

\end{document}